\PassOptionsToPackage{usenames, dvipsnames}{xcolor}
\documentclass[sigconf]{acmart}

\copyrightyear{2018}
\acmYear{2018}
\setcopyright{acmlicensed}
\acmConference[HRI '18]{2018 ACM/IEEE International Conference on Human-Robot Interaction}{March 5--8, 2018}{Chicago, IL, USA}
\acmBooktitle{HRI '18: 2018 ACM/IEEE International Conference on Human-Robot Interaction, March 5--8, 2018, Chicago, IL, USA}
\acmPrice{15.00}
\acmDOI{10.1145/3171221.3171245}
\acmISBN{978-1-4503-4953-6/18/03}

\usepackage{graphicx}
\usepackage[caption=false]{subfig}
\usepackage{amsmath}

\newcommand{\norm}[1]{\left\lVert#1\right\rVert}
\graphicspath{{figures/}}

\usepackage[capitalize]{cleveref}

\colorlet{mylinkcolor}{BrickRed}
\colorlet{mycitecolor}{Green}
\colorlet{myurlcolor}{NavyBlue}

\hypersetup{
  pdfborderstyle={/S/U/W 1},
  linkcolor  = mylinkcolor,
  citecolor  = mycitecolor,
  urlcolor   = myurlcolor,
  colorlinks = true,
}

\usepackage[markup=bfit, deletedmarkup=sout, authormarkup=brackets]{changes}
\definechangesauthor[name={Ale}, color=teal]{AR}
\definechangesauthor[name={Phuong}, color=orange]{PN}
\definechangesauthor[name={Matej}, color=red]{MH}
\definechangesauthor[name={Ugo}, color=brown]{UP}

\begin{document}
\title{Compact Real-time avoidance on a Humanoid Robot for Human-robot Interaction}

\author{Dong Hai Phuong Nguyen}
\affiliation{%
  \institution{Istituto Italiano di Tecnologia}
  \streetaddress{Via Morego 30}
  \city{Genova}
  \country{Italy}
  \postcode{16163}
}
\email{phuong.nguyen@iit.it}

\author{Matej Hoffmann}
\affiliation{%
  \institution{Czech Technical University in Prague}
  \city{Prague}
  \country{Czech Republic}
}
\additionalaffiliation{
  \institution{Istituto Italiano di Tecnologia}
  \streetaddress{Via Morego 30}
  \city{Genova}
  \country{Italy}
  \postcode{16163}
}
\email{matej.hoffmann@fel.cvut.cz}

\author{Alessandro Roncone}
\affiliation{%
  \institution{Social Robotics Lab, Yale University}
  \streetaddress{51 Prospect Street}
  \city{New Haven, CT}
  \country{U.S.A.}
  \postcode{06511}
}
\email{alessandro.roncone@yale.edu}

\author{Ugo Pattacini}
\affiliation{%
  \institution{Istituto Italiano di Tecnologia}
  \streetaddress{Via Morego 30}
  \city{Genova}
  \country{Italy}
  \postcode{16163}
}
\email{ugo.pattacini@iit.it}

\author{Giorgio Metta}
\affiliation{%
  \institution{Istituto Italiano di Tecnologia}
  \streetaddress{Via Morego 30}
  \city{Genova}
  \country{Italy}
  \postcode{16163}
}
\email{giorgio.metta@iit.it}

\renewcommand{\shortauthors}{Phuong D.H. Nguyen et al.}

\begin{abstract}
\makeatletter{}%
With robots leaving factories and entering less controlled domains, possibly sharing the space with humans, safety is paramount and multimodal awareness of the body surface and the surrounding environment is fundamental.
Taking inspiration from peripersonal space representations in humans, we present a framework on a humanoid robot that dynamically maintains such a protective safety zone, composed of the following main components:
(i) a human 2D keypoints estimation pipeline employing a deep learning based algorithm, extended here into 3D using disparity;
(ii) a distributed peripersonal space representation around the robot's body parts;
(iii) a reaching controller that incorporates all obstacles entering the robot's safety zone on the fly into the task.
Pilot experiments demonstrate that an effective safety margin between the robot's and the human's body parts is kept.
The proposed solution is flexible and versatile since the safety zone around individual robot and human body parts can be selectively modulated---here we demonstrate stronger avoidance of the human head compared to rest of the body.
Our system works in real time and is self-contained, with no external sensory equipment and use of onboard cameras only.

\end{abstract}


\maketitle

\makeatletter{}%
\section{Introduction}

A future in which robots can cooperate with each other and with humans requires them to be able to adapt and act autonomously in unstructured and human-populated environments.
In this context, a fundamental issue is safety in human-robot interaction (HRI), where collisions and contacts between the robot and the human are the most important aspect. This is the subject of so-called physical human-robot interaction (pHRI), where the handling of collisions between machines and humans can be divided into two phases: \textsl{pre-impact} and \textsl{post-impact} \cite{de_luca_collision_2006}.

\begin{figure}
  \centering
  \includegraphics[width=.8\linewidth]{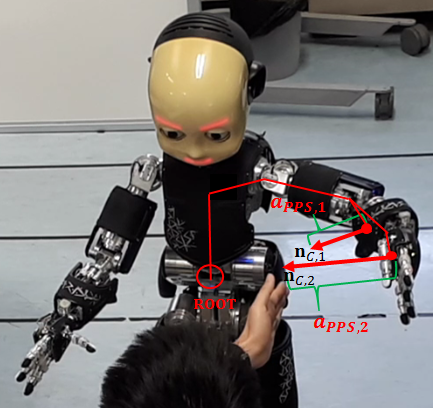}
  \caption{Experimental scenario. The proposed system is able to detect the presence of humans close to the robot's body thanks to a keypoint estimation algorithm combined with a peripersonal space representation. Prior-to-contact activations are translated into a series of distributed control points ($a_{PPS}$ in figure) for \textsl{pre-impact} avoidance. See text for details.}\label{fig:react-ctrl}
\end{figure}

\begin{figure*}
  \centering
  \includegraphics[width=.85\linewidth]{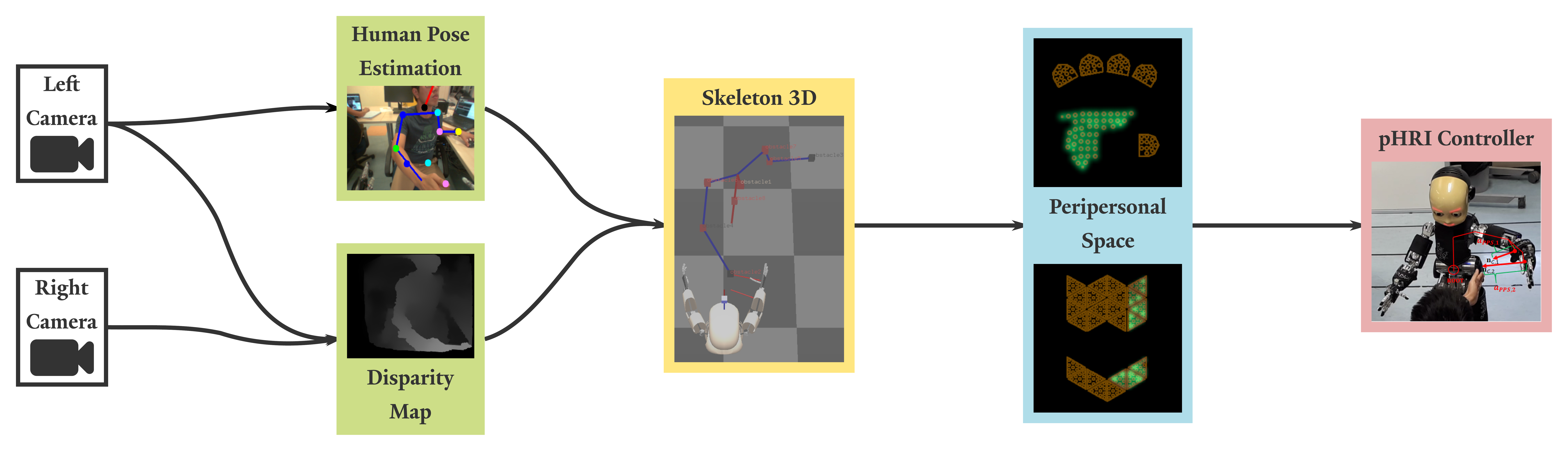}
  \caption{Software architecture for physical human-robot interaction. In this work, we develop a framework composed of: i) a human pose estimation algorithm, ii) a 2D to 3D disparity mapping, iii) a peripersonal space collision predictor, iv) a pHRI robot controller for distributed collision avoidance. See text for details.}\label{fig:pHRI-arch}
\end{figure*}

Research in the \textsl{post-impact} direction relies typically on joint torque or force/torque %
sensors, whose  measurements together with a robot model allow for contact localization (see \cite{haddadin_2017} for a recent survey).
Robot redundancy can then be employed to still accomplish the task while not exerting forces at the obstacle, such as using the residual method \cite{de_luca_collision_2006,de_luca_exploiting_2008}.
Combining the above results with the trajectory scaling strategy, Haddadin et al. \cite{haddadin_collision_2008} present a framework in which the robot switches between different control strategies depending on the collisions detected.
The roots of \textsl{pre-impact} strategies lie in the use of motion planning to find collision-free end-effector trajectories, relying on full knowledge about the robot body and environment. In dynamic scenarios, planning needs to be complemented by reactive strategies such as the potential field approach \cite{khatib_real-time_1990}.
The elastic strips framework \cite{brock_elastic_2002} combines reaching for a goal configuration (global behavior) with reactive obstacle avoidance (local behavior) through incremental modification of a previously planned motion.
Flacco et al.~\cite{flacco_depth_2012} employ a so-called depth-space approach, in which they use an external Kinect sensor and a 2.5D space projection to obtain distances between obstacles and interest points on a robot arm.
These are in turn used to generate corresponding repulsive vectors that are remapped into the joint space, effectively preventing joint movement in the collision direction.
Magnanimo et al.~\cite{magnanimo_2016} do not address the control problem but, unique to their work, they provide a framework to dynamically construct warning and protective safety fields around a manipulator relying on laser scanners to sense the environment and proprioception to sense the manipulator's own velocity.

To guarantee safety in physical human-robot interaction, an essential component is to reliably perceive the human. This usually translates to methods that segment the human body parts from the background and localize them with respect to the robot body.
This is known as human pose / keypoints estimation or skeleton extraction---please refer to recent surveys such as \cite{sarafianos_3d_2016} for 3D pose estimation, and \cite{helten_full-body_2013, ye_survey_2013} for 3D pose estimation from a RGB-D camera.
In a robotics context, particle filters \cite{azad_stereo-based_2007} or Iterative Closest Points (ICP) approaches \cite{droeschel_3d_2011} have been used for human pose estimation.

In this work, we propose a compact, flexible, and biologically inspired solution for safe pHRI in general and collision avoidance in particular that departs from the body of work reviewed above in the following aspects.
First, for human keypoints estimation, no external sensor and no depth sensor is employed. Instead, we present a real-time pipeline that leverages deep learning methods for 2D pose estimation (e.g. \cite{cao_realtime_2016, insafutdinov_deepercut:_2016}) in combination with disparity map computation from a binocular humanoid robot head.
Second, we move significantly beyond solutions that consider the robot's end-effectors exclusively or those where the robot's body is modeled by a set of geometrical collision primitives (e.g., spheres).
Instead, we take inspiration from biology, where the defensive \textsl{peripersonal space} (PPS) representation is maintained by the brain in the form of a network of neurons with visuo-tactile receptive fields (RFs) attached to different body parts and following them as they move (see e.g. \cite{clery_neuronal_2015} for a recent survey). This forms a distributed and much denser coverage of the ``safety margin'' around the whole body. Furthermore, this protective safety zone is dynamically modulated by the state of the agent or by approaching object identity and ``valence'' (positive or negative)---e.g., safety zones around empty vs. full glasses of water \cite{de_haan_influence_2014} or reaction times to spiders vs. butterflies~\cite{de_haan_approaching_2016}.
We capitalize on the PPS representation developed by Roncone et al.~\cite{roncone_IROS_2015,roncone_peripersonal_2016} around the artificial pressure-sensitive skin of the iCub humanoid robot and provide extensions for the purposes of this work.
Lastly, we present a novel robot controller that combines reaching in 3D Cartesian space with simultaneous  obstacle avoidance, with control points created dynamically on the fly on the robot's hands and forearms based on the peripersonal space activations (as illustrated in Fig.~\ref{fig:react-ctrl}).

This article is structured as follows. The next Section discusses the Materials and Methods.
This is followed by the Experiments and Results in \cref{sec:expe-results}, and finally the Conclusion, Discussion, and Future Work in \cref{sec:conclusion}.

\makeatletter{}%
\section{Materials and Methods}
\label{sec:method}
\subsection{The iCub humanoid robot and hardware components utilized}
\label{sec:icub}

The iCub \cite{metta_icub_2010} is a child-sized full humanoid robot (see \cref{fig:react-ctrl}).
For this work, we focus on using its upper body with a 3 degrees of freedom (DoF) torso and two 7 DoF arms. In addition, iCub is equipped with a variety of sensors, of which the cameras, joint encoders, and, indirectly, the artificial skin are relevant here.
The iCub head features a 3 DoF neck and a binocular stereo system composed of 2 identical cameras in a human-like arrangement, with 3 DoFs allowing mechanically coupled tilt motion, and independent version and vergence movements.
Large areas of the iCub body are covered with an artificial electronic skin \cite{maiolino_flexible_2013}, which allows iCub to sense touch---applied pressure.
In this work, the skin is not directly used, but the peripersonal space representation we employ and extend has its origin in the nature of this tactile array.

\subsection{pHRI architecture}\label{sec:pHRI-architecture}

The software architecture of our pHRI framework, presented in \cref{fig:pHRI-arch}, is implemented in C++ and Python with Yarp \cite{fitzpatrick_towards_2008}. Each node in the diagram represents a module in the pipeline.
\begin{itemize}
  \item \textbf{Human pose estimation} processes images from a camera and generates human keypoints, presented in Phase $1$ of \cref{sec:human-estimation};
  \item \textbf{Disparity map} builds the depth map of the environment from images of stereo-camera. This is the result of previous work \cite{pasquale_enabling_2016, fanello_3d_2014};
  \item \textbf{Skeleton3D} constructs the 3D human pose estimation from 2D computations and the depth map of the environment (see Phase $2$ of \cref{sec:human-estimation});
  \item \textbf{Peripersonal Space} serves as a visual-based collision predictor, and is described in \cref{sec:pps};
  \item \textbf{pHRI Ctrl} (physical Human-Robot Interaction Control) translates the spatial perception of robot to motion for safe interaction. The controller is detailed in \cref{sec:avoidance-controller}.
\end{itemize}

\subsection{Human keypoints estimation}\label{sec:human-estimation}

In general, the purpose of human pose estimation algorithms is to provide the configuration of the human body from input image(s) or video. In our case, the input is the set of two images (with resolution of $320 \times 240$ pixels) coming from the iCub cameras.
The iCub head-eye plant differs from common stereo camera systems in that the eyes/cameras are not fixed and move independently in space. This has two consequences:
i) it allows for a compact, self-contained system that does not need any external device to perceive depth information, and
ii) it is not possible to pre-calibrate the plant for the purposes of a disparity map computation--but calibration needs to be performed on the fly.
Because of the latter, we separate our 3D pose estimation algorithm into two successive phases: (i) 2D pose estimation, and (ii) mapping of the 2D pose into the 3D Cartesian space of the robot; they are detailed below.

\subsubsection*{Phase $1$ -- 2D Pose Estimation} The 2D Pose estimation algorithm has the goal of computing the highest probability pixel locations of human keypoints in single camera frames; in our case: head, shoulders, elbows, hands, hips, knees, and ankles.
We denote these locations as \textsl{keypoint pixels}---e.g. $\left[ u_H, v_H \right]$ for the head. The positions of these body parts represent a simplified human model, which is suitable for our task: the keypoints act as obstacles for the robot control algorithm.
We adopt the \textsl{DeeperCut} approach \cite{insafutdinov_deepercut:_2016}, a state-of-art deep learning method for multi-person pose estimation.
The main component of the algorithm is the parts detector constructed by \textsl{ResNet} \cite{he_deep_2016}, a deep learning architecture for object detection.
In addition, DeeperCut uses an incremental optimization approach with integer linear programming \cite{pishchulin_deepcut:_2016} and an image-conditioned pairwise term between body parts to improve the pose estimation quality.
The model is then trained on the \textsl{Leeds Sports Poses} dataset \cite{johnson_clustered_2010} for multiple-person, and the \textsl{MPII Human Pose} dataset \cite{andriluka_2d_2014} for a single person.
The implementation of this algorithm in our system (using Tensorflow \cite{tensorflow_2017} with a single NVIDIA GTX1080i GPU) can provide body parts poses at a frame rate of 30\textsl{ms}. An example of results from this algorithm can be seen in \cref{fig:pps-effect}, panels A.

\subsubsection*{Phase 2 -- 2D to 3D Pose Mapping} The second step of our human keypoints estimation algorithm is to reconstruct 3D poses of body keypoints thanks to the single-camera 2D information from Phase $1$ and a disparity map computed from both cameras (as shown in \cref{fig:pps-effect}, subplot B).
For the reasons detailed above, the disparity map computation does not rely on a pre-existing camera calibration, but needs to rectify both cameras in real-time. For this reason, it is composed of an initial rectification algorithm followed by a disparity estimation step.
The rectification algorithm aligns the two images to a common plane and keeps this transformation up to date with the robot's motion (neck, eyes and torso) \cite{fanello_3d_2014}. The disparity estimation step evaluates pixel displacements between the two rectified images \cite{pasquale_enabling_2016}, making use of the Efficient Large-Scale Stereo (ELAS) Matching algorithm \cite{geiger_efficient_2010}.
The outcome of this 2D to 3D mapping is to complement all the pixels from the left and right cameras with additional depth information in real time.
As a result, it is possible to estimate the 3D Cartesian coordinates of each keypoint pixel estimated in Phase $1$ from the above computed depth---that is, we can compute 3D keypoints coordinates $\left[x_H, y_H, z_H \right]$ from 2D keypoints pixels $\left[u_H, v_H \right]$.
More specifically, we average the estimated 3D positions in the $7 \times 7$ pixels neighborhood of each keypoint to improve robustness.
In addition, we apply biomechanical constraints of human body size and median filters on keypoints' 3D poses to reduce the noise of estimation results. Importantly, this computation is performed in parallel with Phase $1$ thanks to the YARP distributed software architecture.
The human 3D pose estimation that results from this phase is shown in \cref{fig:pps-effect}, panels C.

Moreover, the visual pipeline not only detects the body parts, but identifies them as well.
The recognized body part identities (e.g. head vs. hands) are then exploited to modulate the robots' safety margin and finally regulate robot behavior.
The software developed for this module is freely available \cite{skeleton3d_2017}.

\subsection{Peripersonal space representation}\label{sec:pps}

\begin{figure}
  \includegraphics[width=.6\linewidth]{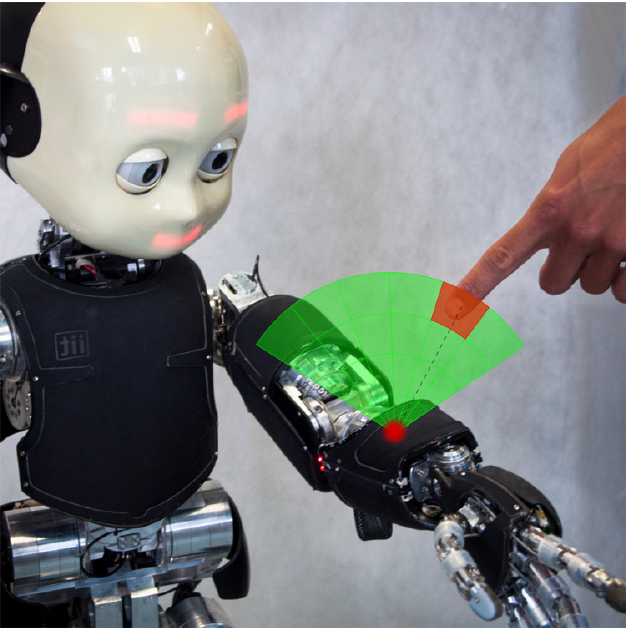}
  \caption{Schematic illustration of PPS receptive fields on the robot. There are five such receptive fields at the palms/hands and $24$ around each of the forearms. From \cite{roncone_IROS_2015} with the permission of authors.}\label{fig:taxelRF-b}
\end{figure}

In Roncone et al. \cite{roncone_IROS_2015,roncone_peripersonal_2016}, a representation of the protective safety zone for the iCub was developed.
Authors chose a distributed representation in which every taxel (tactile element of the robot's artificial skin) is learning a collection of probabilities regarding the likelihood of objects from the environment coming into contact with that particular taxel.
This is achieved by making associations between visual information, as the objects are seen approaching the body, and actual tactile information as the objects eventually physically contact the skin.
A volume was chosen to represent the visual receptive field around every taxel: a spherical sector growing out of every taxel along the normal to the local surface. %
For the purposes of this work, the visual RF size was extended to maximum $45 cm$ away from every taxel%
, motivated by new findings regarding peri-hand PPS \cite{serino_body_2015}; Fig.~\ref{fig:taxelRF-b} schematically illustrates one of such RFs on the robot (in total, there are five RFs on every palm, and $24$ around each forearm).

\begin{figure}
    \centering
    \includegraphics[width=.9\linewidth]{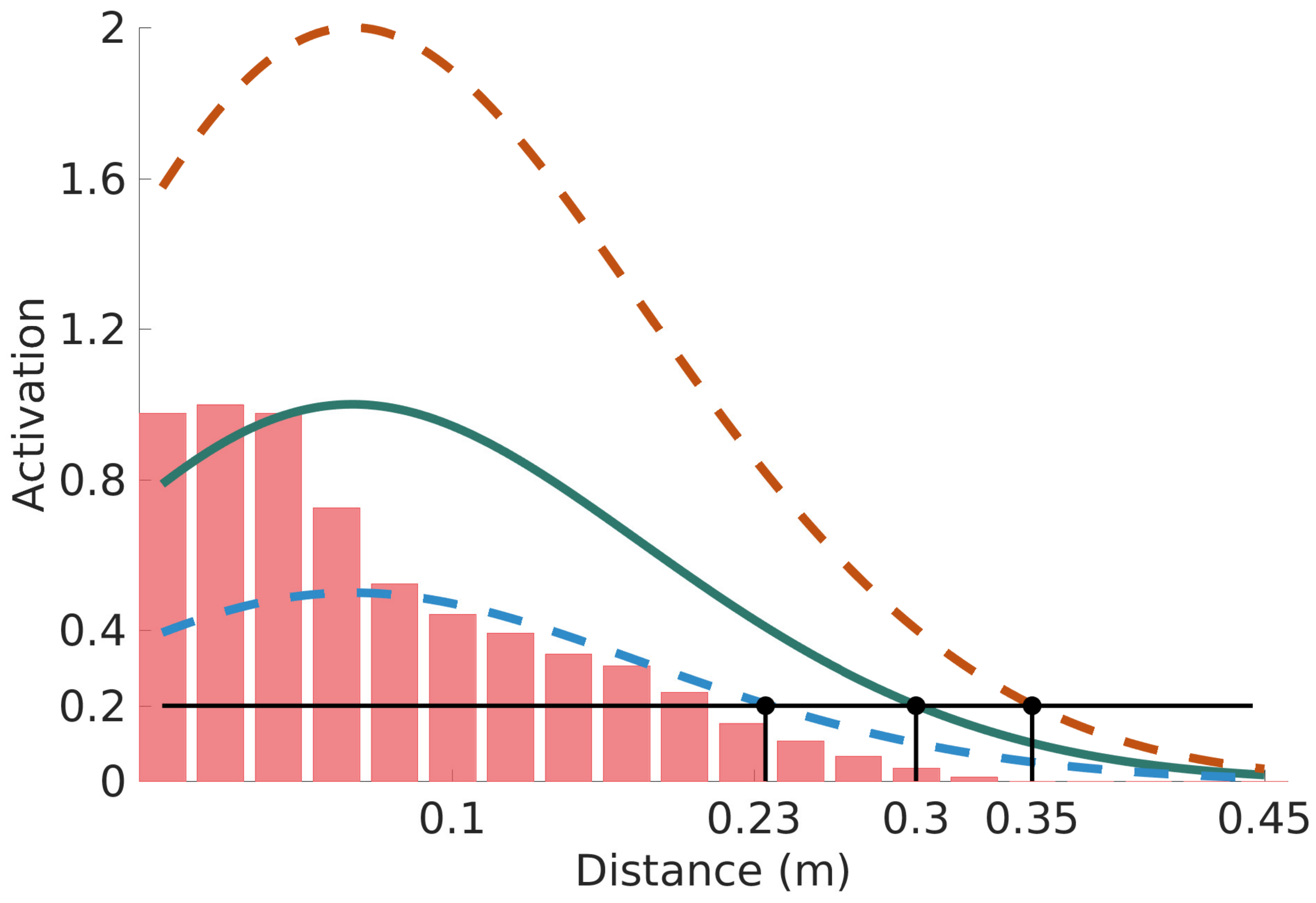}
    \caption{Activation curve of individual taxels' RFs and effect of modulation. Pink bars represent the discretized representation stored into the taxel. The green curve is the result of the Parzen window interpolation technique, whereas the blue and brown dashed curves show the effect of modulation---response attenuation by $50\%$ and positive modulation by $100\%$ respectively. The black line marks an activation threshold of $0.2$, which roughly corresponds to distances from the origin of the RF of $23 cm$, $30 cm$ and $35 cm$ in attenuated, normal, and expanded case respectively.}
    \label{fig:activation-curve-origin}
\end{figure}

Different than in the work of Roncone et al., we did not train the visual RFs here; instead, we designed them uniformly for all taxels.
To preserve compatibility with the original implementation, the taxel RFs have a discrete representation divided in 20 bins that relate distance of stimulus/obstacle to activation, which in turn correspond to the probability of eventual collision---see Fig.~\ref{fig:activation-curve-origin}.
The discrete representation is then interpolated using a Parzen window estimation algorithm, giving rise to the green curve in Fig.~\ref{fig:activation-curve-origin}.
Furthermore, in our implementation, the PPS representation can handle multiple stimuli---objects in the environment---concurrently, with every taxel deriving its response from the closest object.
When the RFs of individual skin taxels are combined, a ``safety margin'' volume around the respective body parts is constructed. Such a ``protective zone'' around the forearm is visualized in Fig.~\ref{fig:pps} (modulated/attenuated version chosen for visualization---see \cref{sec:pps-modulation} below).
The change of the activation w.r.t. the distance (from closest to farthest) is denoted by the change of color from red to light yellow, while the robot body is sketched in gray and black color.

\begin{figure}
    \centering
    \includegraphics[width=\linewidth]{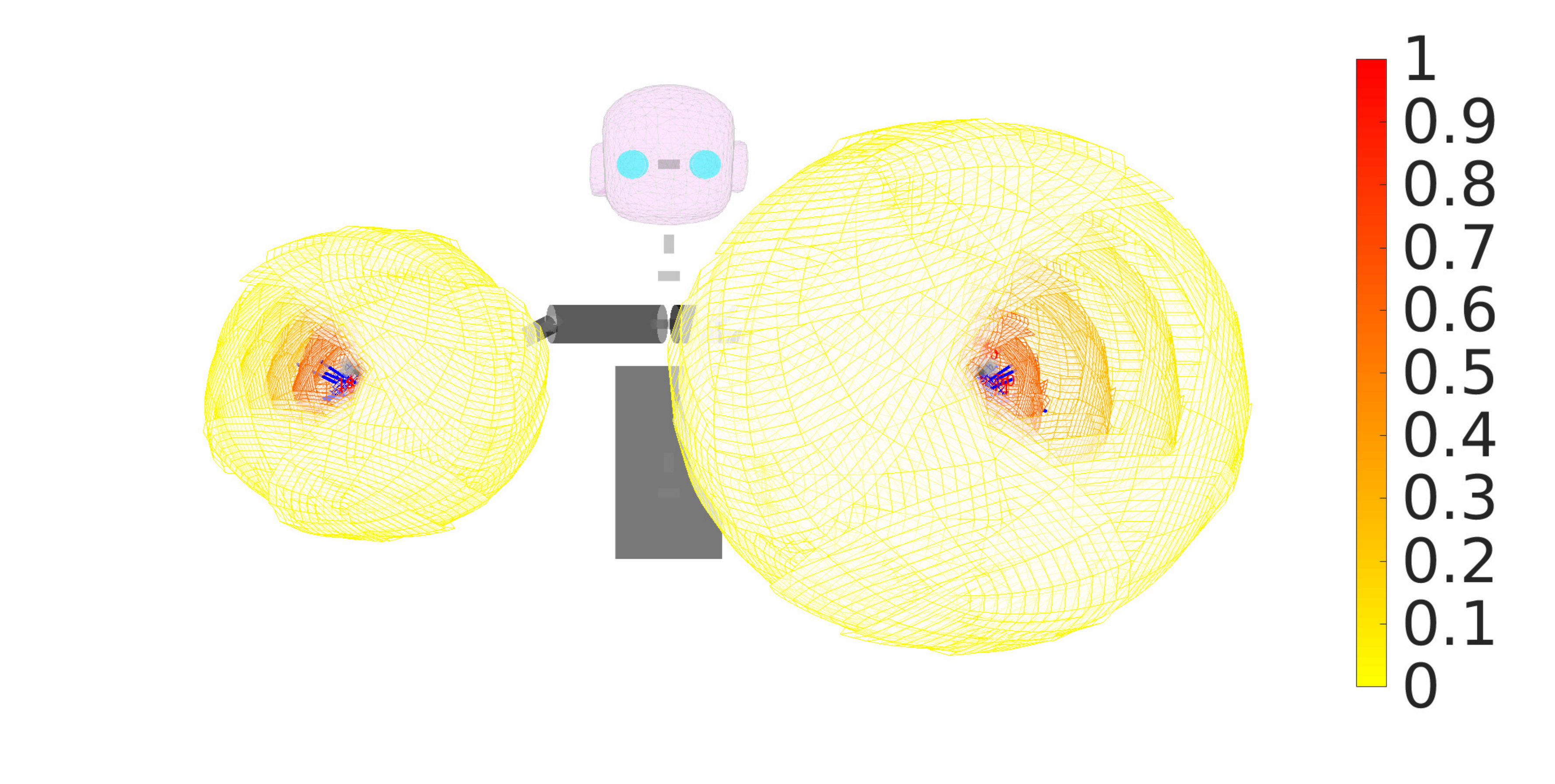}
    \caption{PPS visualization of the two iCub's forearms (front view). In the picture, two negatively modulated / attenuated PPS RFs (75\% for the left forearm and 50\% of the nominal RF for the right forearm).}
    \label{fig:pps}
\end{figure}

\subsubsection*{PPS Modulation}\label{sec:pps-modulation}

\begin{figure}
  \centering
  \subfloat[]{\includegraphics[width=.38\linewidth]{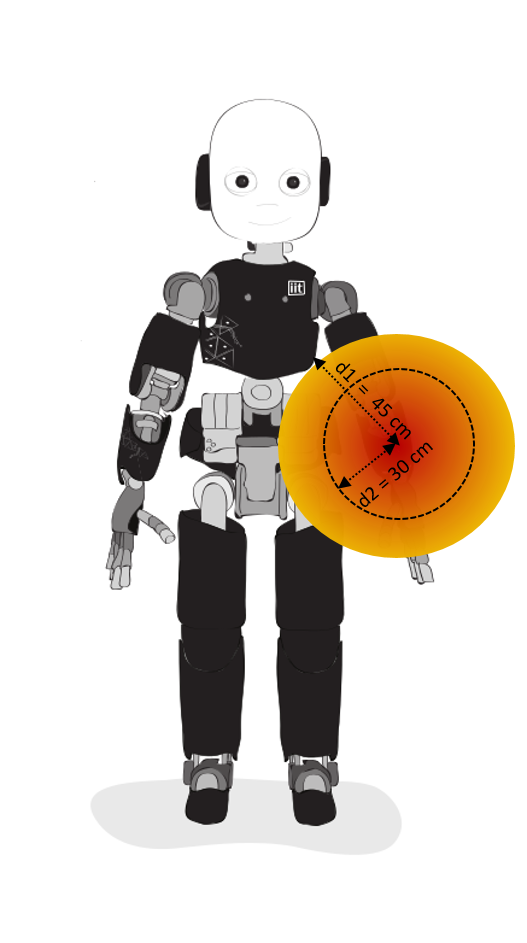}}\qquad
  \subfloat[]{\includegraphics[width=.38\linewidth]{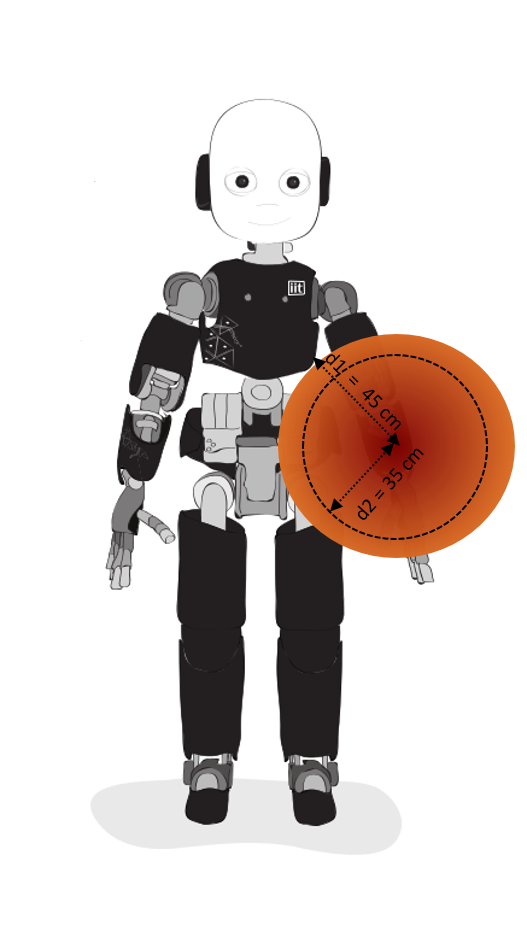}}
  \caption{Peripersonal space modulation illustration. Receptive field extending $45 cm$ and its Gaussian-like distribution of activations (highest at $d = 0 cm$, lowest at the periphery). (a) Nominal RF.  (b) Positively modulated RF.}
  \label{fig:pps_modulation_cartoon}
\end{figure}

Inspired by the human PPS and its modulation reviewed above---e.g., smaller safety zones around empty vs. full glasses of water during reaching \cite{de_haan_influence_2014}---we implement a similar mechanism here.
Such a case is illustrated schematically in Fig.~\ref{fig:pps_modulation_cartoon}: an overall increase / modulation of activation values changes the distance at which a certain activation is reached by an oncoming stimulus, and hence the effective safety margin secured by the robot's responses is also adjusted.
In our case, the modulation will pertain to the ``sensitivity'' of human body parts: for example, while it may be acceptable to come into contact with the hands of the human, the head should be avoided with a much larger safety margin.

We associated a value between $[-1,1]$ to each object, as its ``valence'' $\theta(t)$ with negative values for stimuli where a smaller safety margin is allowed and positive modulation for stimuli that should be avoided with a bigger margin---threatening or fragile objects for example.
The final modulated PPS activation $a_{m,i}(t)$ of the \textsl{i}-th taxel  w.r.t. an object with valence $\theta(t)$ is then calculated as follows:
\begin{equation}
    a_{m,i} (t)= a_i(t)\big[1 +  \theta(t)\big]\quad,
    \label{eq:pps_mod}
\end{equation}

\noindent where $a_i(t)$ is the PPS activation of \textsl{i}-th taxel at instant $t$.
The mechanism is further illustrated in Fig. \ref{fig:activation-curve-origin}. The modulation simply translates the activation curve (y-axis). If associated with a particular activation threshold to trigger behavior (say $a = 0.2$ which will be used here), this will be reached at different distances depending on the modulation---for example at $d2 = 30 cm$ in the nominal case and $d2 = 35 cm$ when subject to positive modulation -- see Fig.~\ref{fig:pps_modulation_cartoon}. This gives rise to effective expansion/shrinking of the aggregated safety margin (composed of multiple RFs), as shown in Fig.~\ref{fig:pps}.
The software developed for this module is freely available online \cite{peripersonal-space_2017}.

\subsection{Reaching with avoidance on the whole arm surface}\label{sec:avoidance-controller}

The proposed controller has its roots in the Cartesian controller of Pattacini et al.~\cite{pattacini_experimental_2010} who proposed an inverse kinematics solver and minimum-jerk controller for the iCub robot.
There, the solver is formulated as a nonlinear constrained optimization problem expressed in the joint position space and makes use of the IpOpt library \cite{wachter_implementation_2006}; it is decoupled from the controller part.
In this work, we propose a solution that unifies the inverse kinematics and the robot control problems into a single formulation: the problem is directly expressed in the joint velocity space and the solutions to it---joint velocities---can be directly used to control the robot.
More specifically, at every time step $t = \bar{t}$ (with a period $T_S = 20 ms$), we compute the desired joint velocities $\dot{\mathbf{q}}^* (\bar{t})$ by solving the following:
  \begin{equation}
    \renewcommand{\norm}[1]{\left\lVert#1\right\rVert}
    \begin{split}
    \dot{\mathbf{q}}^* &= \underset{\dot{q}\in \mathbb{R}^n}{arg\,min} \Big[\norm{\mathbf{x}_{EEd}-\bigl(\bar{\mathbf{x}}_{EE}+T_S\mathbf{J}(\bar{\mathbf{q}})\dot{\mathbf{q}}\bigr)}^2 \Big]\\
    & \mathbf{s.t.} %
        \left.
        \begin{cases}
          &\mathbf{q}_L < \bar{\mathbf{q}}+T_S\dot{\mathbf{q}} < \mathbf{q}_U\\
          &\dot{\mathbf{q}}_L < \dot{\mathbf{q}} < \dot{\mathbf{q}}_U\\
        \end{cases}
        \right.
    \end{split}
    \label{eq:reactCtrl_optimization}
  \end{equation}

\noindent where $\bar{\mathbf{q}} = \mathbf{q} (\bar{t})$ represents the instantaneous configuration of the $n$ joints of the robot arm ($n=7$ in this work), $\mathbf{\bar{x}}_{{EE}} = \mathbf{x}_{{EE}} (\bar{t})$ is the 6D pose of the end-effector in the Cartesian space (comprising of position and orientation), $\mathbf{x}_{EEd}$ is the desired end-effector 6D pose, and $\mathbf{J}(\bar{\mathbf{q}})$ is the Jacobian.
Equation \ref{eq:reactCtrl_optimization} models the reaching task as an optimization problem, specifically as a minimization of the distance between the desired end-effector pose $\bar{\mathbf{x}}_{EEd}$ and one-step-ahead prediction giving the future pose that can be computed from the current pose $\bar{\mathbf{x}}_{EE}$, current joint configuration $\mathbf{q} (\bar{t})$, and joint velocities $\dot{\mathbf{q}}$ (the unknown).
It is well established that, given a sufficiently small $T_S$, this can be approximated by the Jacobian map $\mathbf{J}(\bar{\mathbf{q}})$ multiplied by the vector of joint velocities $\dot{\mathbf{q}}$, which are the unknown of the problem.
To ensure feasibility of the optimal solution, the minimization needs to be carried out under a set of constraints that confine the one-step-ahead estimated joint position $\bar{\mathbf{q}}+T_S\dot{\mathbf{q}}$ within the feasible joint range $\left[ \mathbf{q}_L, \mathbf{q}_U \right]$ (first row); furthermore, we limit the estimated joint velocity $\dot{\mathbf{q}}$ to be within the maximum and minimum limits $\left[ \dot{\mathbf{q}}_L, \dot{\mathbf{q}}_U \right]$ as per specifications of the respective robot actuators (second row). Other non-linear constraints can be conveniently added for further specialization of the control loop (see below).
Again, we make use of the Ipopt library \cite{wachter_implementation_2006}.

The reaching task has to be reconciled with simultaneous obstacle avoidance. To compute the obstacles, we capitalize on the PPS representation described in \cref{sec:pps}. Roncone et al. \cite{roncone_IROS_2015,roncone_peripersonal_2016} aggregate the distributed PPS activations into a single locus and strength per body part (forearm or hand), using a weighted average of position $\mathbf{P}_C$, normal direction $\mathbf{n}_C$ (to the skin at the individual taxels), and activation $\mathbf{a}_{PPS}$  as follows:
  \begin{equation}
    \begin{split}
      \mathbf{P}_C(t) &= \frac{1}{k}\sum^k_{i=1}\big[a_i(t) \cdot \mathbf{p}_i(t)\big]\\
      \mathbf{n}_C(t) &= \frac{1}{k}\sum^k_{i=1}\big[a_i(t) \cdot \mathbf{n}_i(t)\big]\\
      \mathbf{a}_{PPS}(t) &= \overset{k}{\underset{i=1}{max}}\big[ a_i(t) \big]
    \end{split}
  \end{equation}

\noindent with subscript \textsl{i} denoting the \textsl{i}-th taxel, $i = 1 \dots k$. The idea of PPS activation aggregation is illustrated in Fig.~\ref{fig:pps-effect-origin} where the high-resolution activations on forearm and hand (panels 3 and 4) are combined into single vectors per body part -- red arrows in panel C. These aggregated vectors acting along the normal are schematically illustrated in Fig.~\ref{fig:react-ctrl}.
The weighted average position $\mathbf{P}_C$ is employed as a new control point $C_i$ that can then be used to bring about ``reaching'' or avoidance behaviors  along the normal $\mathbf{n}_C$.
However, only one task---either reaching or avoidance---at a time can be accomplished for a single control point.
In this work, we introduce a novel solution that enables reaching with simultaneous obstacle avoidance by incorporating these additional control points  into the controller described in Equation \ref{eq:reactCtrl_optimization} as additional joint velocity constraints.
This remapping of the Cartesian ``repulsive vectors'' into joint space constraints is described by the following equations:
  \begin{equation}
    \begin{split}
      \mathbf{s} &= - \mathbf{J}^T_C \cdot \mathbf{n}_C \cdot V_C \cdot a_{PPS}\\
      \dot{\mathbf{q}}_{L,j} &= max \big\{ V_{L,j}, s_j\big\}\ ,\quad s_j \geq 0\\
      \dot{\mathbf{q}}_{U,j} &= min \big\{ V_{U,j}, s_j\big\}\ ,\quad s_j<0
    \end{split}
  \label{eq:velocity_constraints}
  \end{equation}

\noindent where $C$ is a control point belonging to a generic robot link, $\mathbf{J}_C$ is its associated Jacobian, $V_C$ is a gain factor for avoidance, and $V_L, V_U$ are a predefined set of bounding values of joint velocity, e.g. $\pm 25 deg/s$.
When projecting repulsive vectors in joint space, we obtain the value $s_j$, whose component $\mathbf{s}$ represents the ``degree of influence'' of the Cartesian constraint on the \textsl{j}-th joint.
From these, the admissible upper ($\dot{\mathbf{q}}_U$) and lower ($\dot{\mathbf{q}}_L$) velocity limits of those joints influenced by the risk of collision are reshaped.
Differently than \cite{flacco_depth_2012}, which inspired our approach and where joints that would move toward the obstacle are stopped, we bring about an active avoidance behavior. The avoidance action is thus proportional to the ``threat level'', $a_{PPS}$, and for individual joints to how much each joint can contribute in the current configuration.
To improve smoothness, the desired target velocities are fed to a minimum-jerk filter; velocities are then integrated to compute target joint positions, which are directly fed to low-level position-direct motor controllers.
This last step is standard to most robotic platforms---see \cite{pattacini_experimental_2010} for details on this for what concerns the iCub.

The proposed approach is relevant to the robot control community in that it uses a constrained non-linear optimization technique for inverse kinematics and control.
By sidestepping the computation of an analytical solution to inverse kinematics, the system is automatically immune from singularities; however, it may incur in sub-optimal local minima. Yet, this problem is mitigated in practice due to the large number of degrees of freedom available.
In future work, the framework can be complemented by Cartesian planning algorithms (cf. \cref{sec:conclusion}).
To our knowledge, this is the first attempt at developing a robot control software that yields velocity profiles adopting nonlinear optimization.
In addition, the novelty of this approach has been further enhanced by the integration of avoidance capabilities.
Software developed for this module is also freely available online \cite{react-control_2017}.

\makeatletter{}%
\section{Experiments and Results}\label{sec:expe-results}

\begin{figure}
  \subfloat[Normal (unmodulated) PPS. Human hand triggers high activation of the right palm and the inner part of the right forearm PPS.]{\includegraphics[width=.95\linewidth]{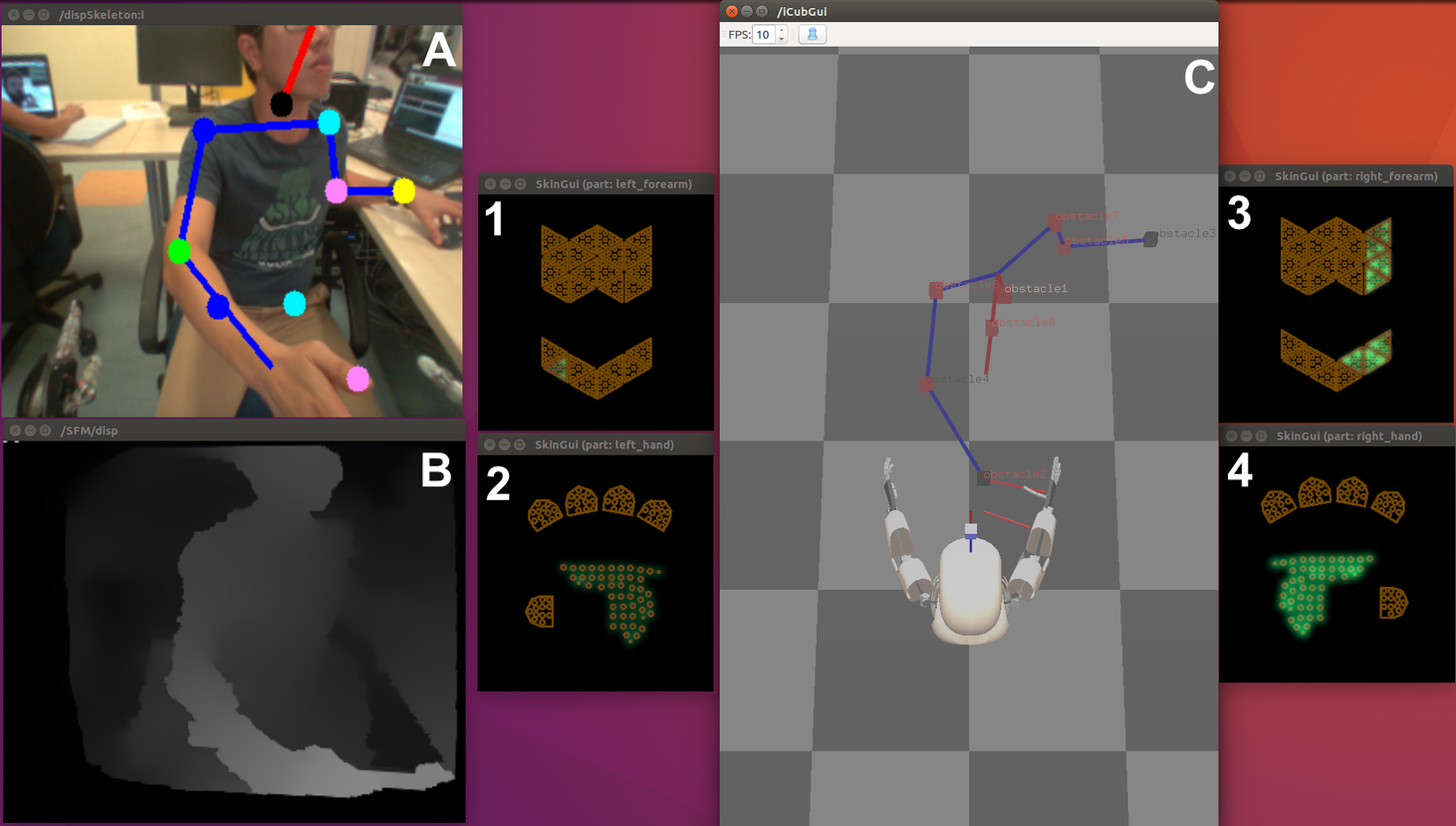} \label{fig:pps-effect-origin}}

  \subfloat[Modulated PPS with attenuated response for hands and arms.]{\includegraphics[width=.95\linewidth]{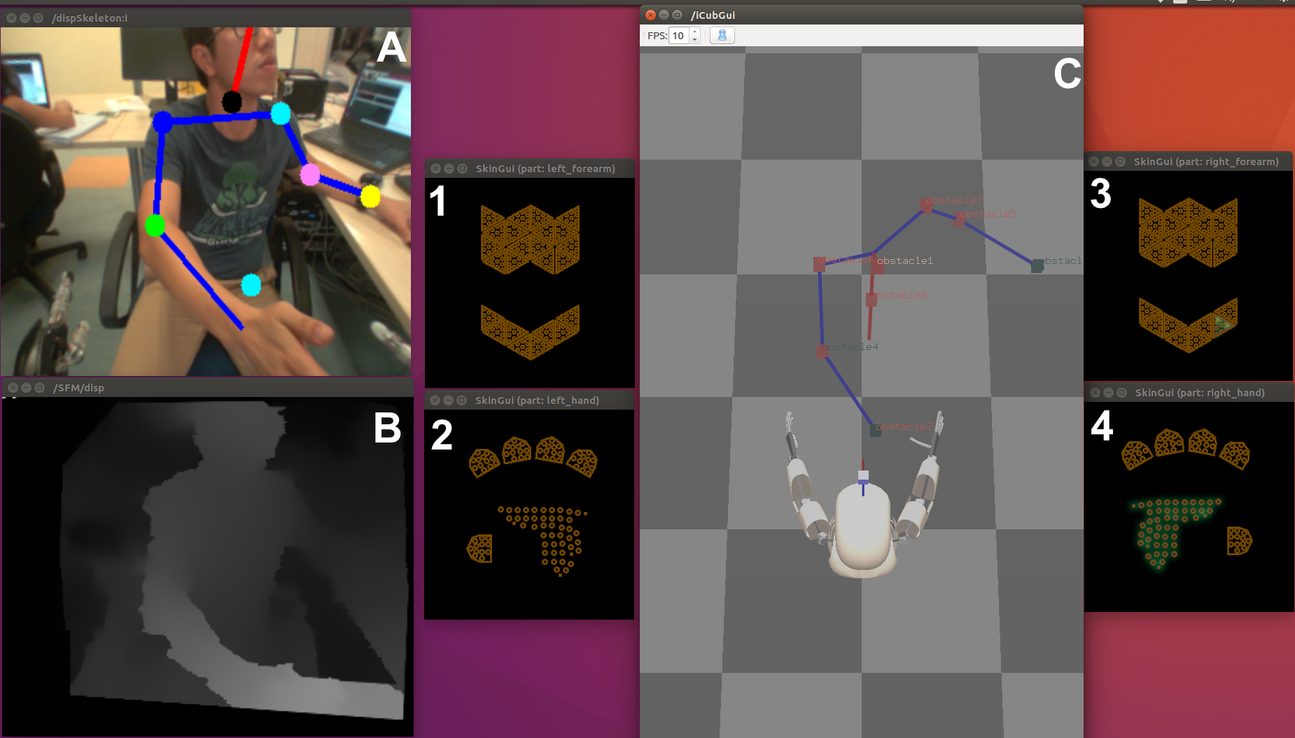} \label{fig:pps-effect-modulation}}

  \caption{Perception, PPS representation, and its modulation on iCub during interaction with human.
  Panels 1,2,3,4: PPS activations on left forearm, left hand, right forearm, and right hand, respectively visualized using iCub $skinGuis$. Taxels turning green express the activation of the corresponding PPS representation (proportional to the saliency of the green).
  Panel A: skeleton of human in 2D.
  Panel B: disparity map from stereo-vision.
  Panel C: estimated skeleton of human in 3D alongside with the iCub robot. Red arrows in this panel show the direction and magnitude of aggregated PPS activations on iCub body parts w.r.t. the obstacle (human right hand).}\label{fig:pps-effect}
\end{figure}

In this section, we describe experimental results regarding three different HRI scenarios, in which the robot executes pre-defined tasks (reaching for a position, following a trajectory) and the human experimenter interferes.
For the purposes of this work, we perform pilot experiments in which trained human participants interact with the robot; these prototypical experiments (described in \cref{sec:reaching-normal,sec:reaching-dif,sec:reaching-circle}) lay the ground for future user studies, which are out of the scope of this work.
In all experiments, a minimum threshold of $\textbf{a}_{PPS} = 0.2$ is set for the avoidance behavior to be triggered; also, the robot is commanded to either static or moving position targets, with a fixed orientation (palms pointing inwards).
We report results using one arm of the robot with PPS around its hand and forearm and control of 7 joints of the arm; however, the framework operates in the same way for both arms and three torso joints could be toggled on using the very same controller.
During robot operation, data from several software modules (\textsl{skeleton3D}, \textsl{Peripersonal Space}, \textsl{pHRI Ctrl}) and robot sensors (joints encoders, cameras) are recorded and later analyzed in Matlab.
\cref{fig:pps-effect} provides a static overview of the perception part of the pipeline.
Please refer to the accompanying video for an overview of the setup and a qualitative evaluation of the performance: \url{https://youtu.be/A9Por3anPJ8}.
Our framework has been released under the LGPL v2.1 open-source license, and is freely accessible on GitHub \cite{skeleton3d_2017,peripersonal-space_2017,react-control_2017}; the control architecture is readily available for any iCub robot, and can be extended to other platforms.

\subsection{Reaching for static target with simultaneous avoidance}\label{sec:reaching-normal}

\begin{figure}
  \centering
  \includegraphics[width=\linewidth]{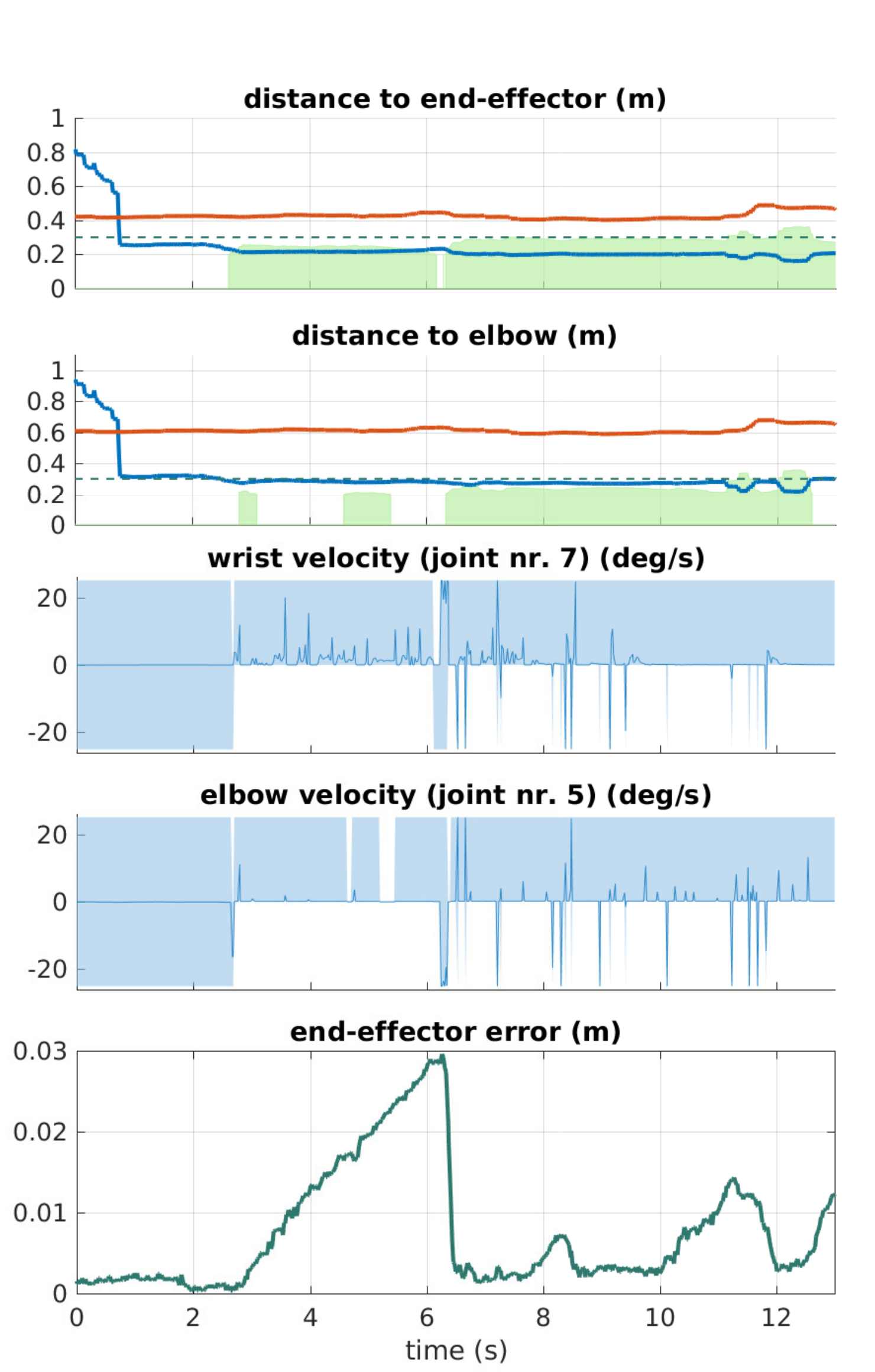}  %
  \caption{Reaching for static target with avoidance. Top two panels (\textsl{Distance-Activation}) for robot end-effector / elbow: blue and orange lines: distance from left hand and head of human respectively; light green areas: PPS activations $\textbf{a}_{PPS}$ on robot body parts (hand -- top panel, forearm -- 2nd panel); green dotted line: distance at which PPS activation exceeds $0.2$ and avoidance is activated (cf. \cref{fig:activation-curve-origin});
  Panels 3-4 (\textsl{Joint velocity}):
  Joints velocity (in blue) and their adaptive bounds (light blue band) -- two selected joints only. Bottom panel (\textsl{End-effector error}): Euclidean distance between reference and actual position of the end-effector.}
  \label{fig:exp_all_normal}
\end{figure}

In this experiment, the iCub is tasked with maintaining its end-effector at a predefined position (i.e. the control target is a static 3D point), while avoiding collisions when the human is approaching the robot body.
Note that collision avoidance has always priority, since it is a constraint for the controller and needs to be satisfied at all times, whereas the reaching task is expressed as a criterion to be minimized.
That is, when the human interferes, the robot should be able to avoid contact with the human at any given moment, departing from its predefined static target when necessary.
Results from this experiment are shown in \cref{fig:exp_all_normal}.
The human body parts activate the PPS  when they enter their RFs ($45 cm$ zone from the skin surface), and increase the activations if they continue to get closer to the robot's arm.
However, there is no effect on joint velocities until the activations reach the threshold of $0.2$, which is corresponding to approximately $30 cm$ away from the skin surface (shown by the dashed green straight line in top two panels; cf. Fig.~\ref{fig:activation-curve-origin}). The end-effector error is minimal there.
After about $2.7s$ into the experiment, the human body parts induce super-threshold PPS activations at the robot hand and partially at the forearm.
This propagates into the robot control algorithm that adaptively tunes the joint velocity limits for all affected joints, as specified by \cref{eq:velocity_constraints}. As shown in \cref{fig:exp_all_normal}, panels $3$ and $4$ (only two joints out of seven are shown for clarity), the range of velocity limits is reduced and the joint velocities are consequently constrained such that avoidance is generated.
The activations on robot's left hand influence all the joints on the chain, while forearm activations only those from the elbow up (more proximal).
Eventually, the robot's end-effector cannot stay at the desired position but avoids the human body parts when they approach, shown by the growing error in the bottom panel of \cref{fig:exp_all_normal} (e.g. after $2.7s$).

\subsection{Reaching with modulation for different body parts (human head vs. hand)}\label{sec:reaching-dif}

\begin{figure}
  \centering
  \includegraphics[width=\linewidth]{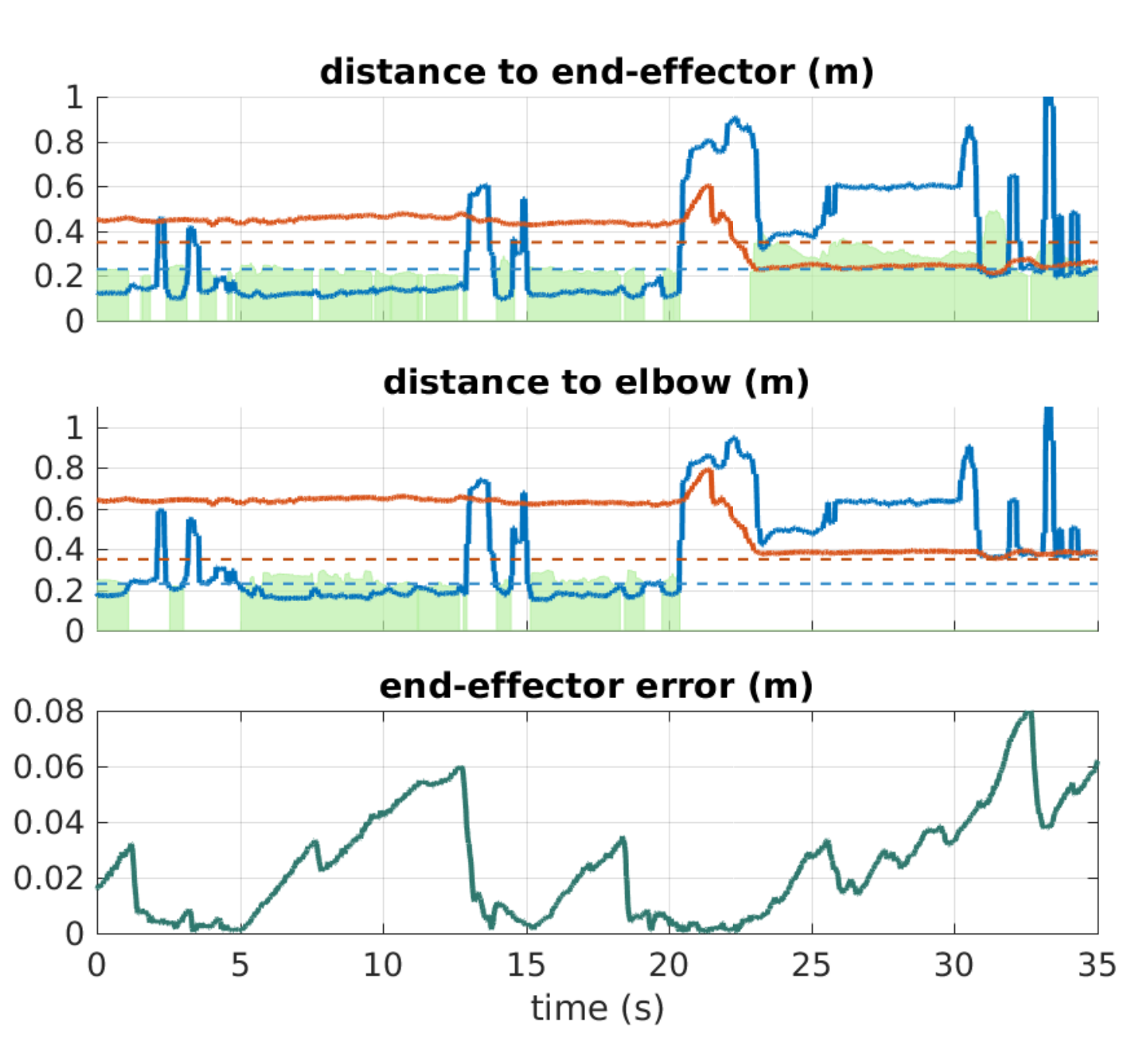} %
  \caption{Static target and PPS modulation: human hand vs. head. See \cref{fig:exp_all_normal} for explanation of individual panels and text for details.}\label{fig:exp_all_dif}
\end{figure}

The setup for this experiment is similar to \cref{sec:reaching-normal}. The main difference is that we assigned different relevance for the different human body parts, as specified in \cref{sec:pps-modulation}.
This directly relates to a realistic human-robot interaction in which the safety of some body parts (e.g., head) should be guaranteed with a bigger margin than for others (e.g., arms).
To illustrate the principle, we apply a $50\%$ PPS attenuation at the hands (i.e. $\theta = -0.5$ in Eq.~\ref{eq:pps_mod}; see also blue dashed curve in Fig.~\ref{fig:activation-curve-origin} and left forearm PPS in Fig.~\ref{fig:pps}), while we positively modulate the PPS pertaining to human head (valence $1.0$; red dashed curve in Fig.~\ref{fig:activation-curve-origin}).
A potential interaction scenario that can take advantage of PPS modulation is physical human robot cooperation in a shared environment, where the robot may need to come into contact with the human hands to receive or hand-over objects (active contacts), but must always avoid any collisions with her head.
Results from the experiment are reported in \cref{fig:exp_all_dif} and structured similarly to \cref{sec:reaching-normal}, with the exception that we do not report joint velocity plots for clarity.
Due to the reduced safety margin, the human left hand (blue line in panels $1$ and $2$ of \cref{fig:exp_all_dif}) can get closer to the robot's end-effector and elbow respectively, while it only activates the robot's PPS slightly, just above $0.2$ (before $t \simeq 5 s$).
As a consequence of this, there are only small regulations applied on joint velocity bounds, and the robot can still perform the task successfully (as shown by the small error in panel $3$ of \cref{fig:exp_all_dif}, $t = [0 s\ \dots\ 5 s]$).
At $t \simeq 22.5 s$, the human head enters the peripersonal space of the end-effector and triggers a strong response of the PPS representation.
Therefore, in order to preserve safety, the robot cannot maintain the reaching task (the end-effector error in panel $3$ increases) but is successful in maintaining a safe margin from the human head.

\subsection{Following a circle while avoiding human}\label{sec:reaching-circle}

\begin{figure}
  \centering
  \includegraphics[width=\linewidth]{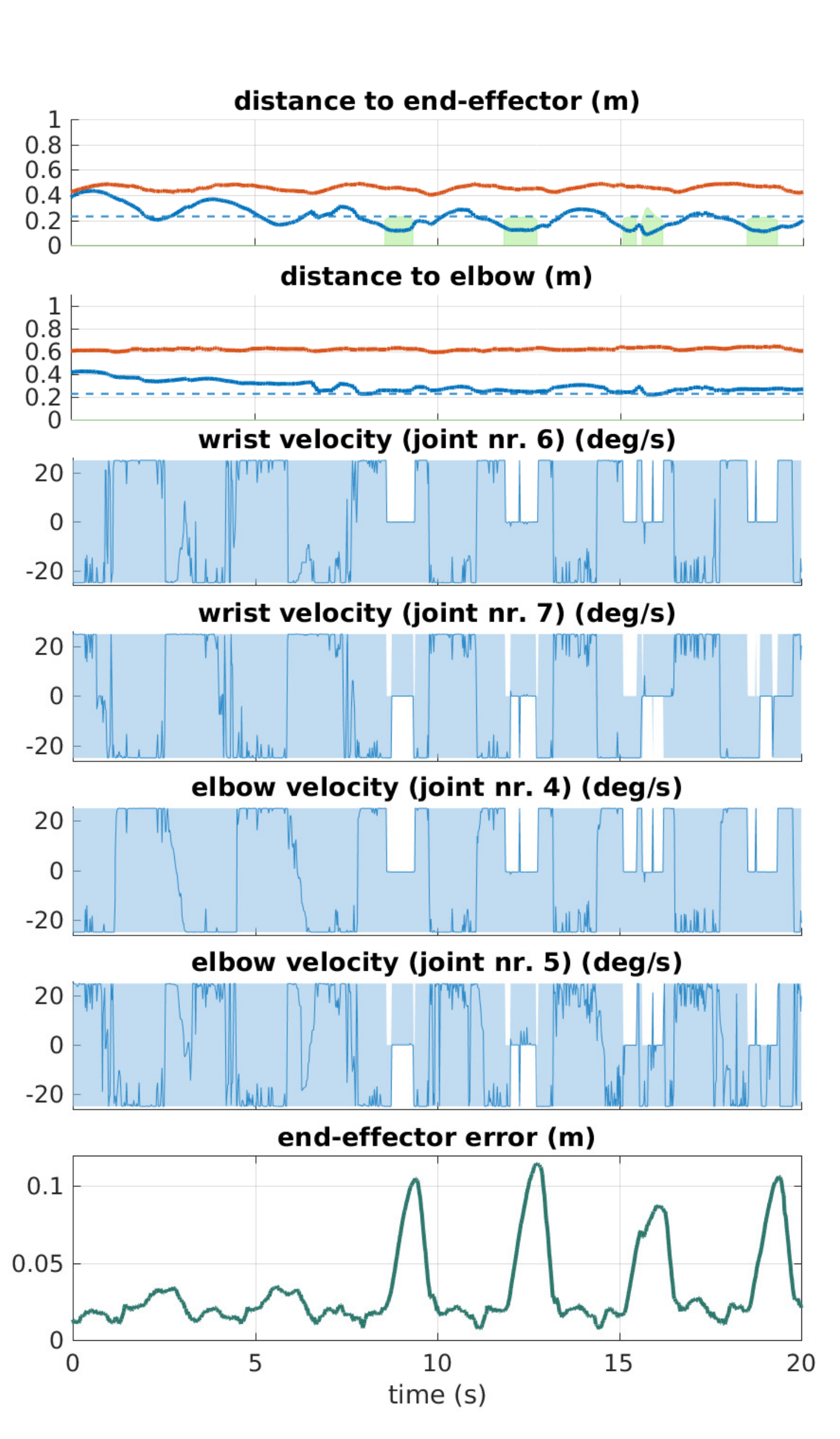} %
  \caption{Moving target on a circle.  See \cref{fig:exp_all_normal} for explanation of individual panels and text for details.}\label{fig:exp_all_circle}
\end{figure}

In this experiment, the robot is commanded to follow a circular trajectory with the left arm, while the human interferes with this task, hence triggering the avoidance behavior. The valences of human body parts are kept same as in \cref{sec:reaching-dif} (attenuation for hand; boosting for head).
Results are shown in \cref{fig:exp_all_circle}, with 4 joint velocity subplots (2 for the elbow and 2 for the wrist).
Similar to the static reaching case, when the human parts approach close enough to the robot's arm ($t \simeq 8 s$), the controller chooses to avoid the human rather than continuing to follow the desired path.
This behavior can be recognized by the relation between distances-activations (in the \textsl{Distance-Activation} panels in \cref{fig:exp_all_circle}) and the changes of joint velocity bounds (in panels $3$ to $6$).
Without the interference of the human (e.g. before $t = 8 s$), the bounds of joint velocities keep the preset values ($\pm25\textsl{deg/s}$), leading to successful tracking behavior of the robot's end-effector on the desired trajectory (small error shown in the panel $7$).
Conversely, the joint velocity bounds are dynamically adapted when the human approaches (the blue band reduces), thus causing the robot to deviate from the demanded trajectory (error increases cyclically after $t \simeq 8 s$).

\makeatletter{}%
\section{Discussion and Future work} \label{sec:conclusion}

We developed and tested a new framework for safe interaction of a robot with a human composed of the following main components:
(i) a human 2D keypoints estimation pipeline employing a deep learning based algorithm, extended here into 3D using disparity;
(ii) a distributed peripersonal space representation around the robot's body parts;
(iii) a new reaching controller that incorporates all obstacles entering the robot's protective safety zone on the fly into the task.
The main novelty lies in the formation of the protective safety margin around the robot's body parts---in a distributed fashion and adhering closely to the robot structure---and its use in a reaching controller that dynamically incorporates threats in its peripersonal space into the task.
The framework was tested in real experiments that reveal the effectiveness of this approach in protecting both human and robot against collisions during the interaction.
Our solution is compact and self-contained (onboard stereo cameras in the robot's head being the only sensor) and flexible, as different modulations of the defensive peripersonal space are possible---here we demonstrate stronger avoidance of the human head compared to rest of the body.

Relying solely on visual perception of the human is, however, not enough to warrant safe interaction under all circumstances.
Additional safety layers would naturally fall into the \textsl{post-impact} phase. In our robot, this could be contacts perceived on the artificial skin or from the force/torque sensors located in the upper arm.
Such contacts can be seamlessly integrated into the controller presented here, making the whole framework multimodal and more robust. At the same time, the proposed solution is not restricted to the iCub humanoid robot and its adaptation to other platforms (with RGB-D sensors instead of stereo cameras; without artificial skin; with a different number of DoF etc.) would be  straightforward.
The ``pHRI controller'' presented here is unique in that it combines a local inverse kinematics solver with a controller in a single module. In the future, we are planning to further extend it to enable processing of multiple targets in Cartesian space---for different control points on the robot body---and to couple it with a global whole-body planner (e.g., \cite{nguyen_fast_2016}).
Finally, not only static distances between the robot and human can be considered, but both human and robot velocities could be taken into account (as dealt with by \cite{roncone_peripersonal_2016} and \cite{magnanimo_2016} respectively).

\begin{acks}
  D.H.P.N. was supported by a Marie Curie Early Stage Researcher Fellowship (H2020-MSCA-ITA, SECURE 642667).
  M. H. was supported by the Czech Science Foundation under Project GA17-15697Y.
  A. R. was supported by a subcontract with Rensselaer Polytechnic Institute (RPI) by the Office of Naval Research, under Science and Technology: Apprentice Agents.
  We would also like to thank Jordi Ysard Puigbo for discussions on the peripersonal space modulation.
\end{acks}

\bibliographystyle{ACM-Reference-Format-nosorted}	%
\newcommand{\showDOI}[1]{\unskip}
\newcommand{\showURL}[1]{\unskip}
\bibliography{bibliography}

\end{document}